\documentclass[10pt]{article} 

\usepackage[preprint]{tmlr}

\usepackage{amsmath,amsfonts,bm}

\def\eqref#1{equation~\ref{#1}}

\def\1{\bm{1}}

\DeclareMathAlphabet{\mathsfit}{\encodingdefault}{\sfdefault}{m}{sl}
\SetMathAlphabet{\mathsfit}{bold}{\encodingdefault}{\sfdefault}{bx}{n}

\usepackage{hyperref}
\usepackage{graphicx}%
\usepackage{multirow}%
\usepackage{amsmath,amssymb}%
\usepackage{mathrsfs}%
\usepackage[title]{appendix}%
\usepackage{xcolor}%
\usepackage{textcomp}%
\usepackage{manyfoot}%
\usepackage{booktabs}%
\usepackage{algorithmicx}%
\usepackage{listings}%
\usepackage{latexsym}
\usepackage{url}
\usepackage{algcompatible}
\usepackage{algpseudocode}
\usepackage{amssymb}
\usepackage{multirow}
\usepackage{verbatim}
\usepackage{scrextend}
\usepackage{float}
\usepackage{subfig}
\usepackage{array}
\usepackage{wrapfig}
\usepackage{mathtools}

\usepackage{tablefootnote}

\newcolumntype{C}[1]{>{\centering\let\newline\\\arraybackslash\hspace{0pt}}m{#1}}

\usepackage{tikz}
\usetikzlibrary{shapes,arrows}
\usepackage{amsmath,bm}
\usepackage{verbatim}
\usepackage{subfig}

\usepackage{enumitem}
\usepackage{xcolor,colortbl}
\usepackage{booktabs} 
\usepackage{arydshln} 

\usepackage{float}
\usetikzlibrary{positioning}
\usepackage{tkz-graph}

\usetikzlibrary{decorations.pathreplacing,intersections}
\usepackage{wrapfig}
\usepackage{pbox}
\usepackage{colortbl}
\usepackage{xcolor,soul,framed}
\colorlet{shadecolor}{yellow}
\usepackage{eqparbox}
\usepackage{url}
\usepackage{pifont}
\usepackage{amssymb}
\usepackage{amssymb}
\usepackage{amsmath}
\usepackage{graphics}
\usepackage{graphicx}

\usepackage{algorithm2e}

\usepackage{booktabs, multirow} 
\usepackage{soul}
\usepackage{colortbl}
\usepackage{subfig}
\usepackage{enumitem}
\usepackage{tabularray}
\usepackage{todonotes}
\usepackage{arydshln}
\usepackage[export]{adjustbox}
\newcommand{\modelname}{{\texttt{PARADOX}}}
\definecolor{maroon}{cmyk}{0,0.87,0.68,0.32}

\author{Ayan Sengupta \email ayan.sengupta@ee.iitd.ac.in \\
       \addr Department of Electrical Engineering\\
       Indian Institute of Technology Delhi\\
       \AND
       Md. Shad Akhtar \email shad.akhtar@iiitd.ac.in \\
       \addr Department of Computer Science \& Engineering\\
       Indraprastha Institute of Information Technology Delhi\\
       \AND
       Tanmoy Chakraborty\thanks{Corresponding Author}  \email tanchak@iitd.ac.in \\
       \addr Department of Electrical Engineering\\
       Yardi School of Artificial Intelligence\\
       Indian Institute of Technology Delhi\\}

\def\month{10}  
\def\year{2024} 
\def\openreview{\url{https://openreview.net/forum?id=fzP4qIiVIh}} 

\RestyleAlgo{ruled}
\begin{document}

\title{Persona-aware Generative Model for Code-mixed Language}

\maketitle

\begin{abstract}
Code-mixing and script-mixing are prevalent across online social networks and multilingual societies. However, a user's preference toward code-mixing depends on the socioeconomic status, demographics of the user, and the local context, which existing generative models tend to ignore while generating code-mixed texts. In this work, we make a pioneering attempt to develop a persona-aware generative model to generate texts resembling real-life code-mixed texts of individuals. We propose {\modelname}, a persona-aware generative model for code-mixed text generation, which is a novel Transformer-based encoder-decoder model that encodes an utterance conditioned on a user's persona and generates code-mixed texts {\em without} monolingual reference data. We propose an alignment module that re-calibrates the generated sequence to resemble real-life code-mixed texts. {\modelname} generates code-mixed texts that are semantically more meaningful and linguistically more valid. To evaluate the personification capabilities of \modelname, we propose four new metrics -- CM BLEU, CM Rouge-1, CM Rouge-L and CM KS. On average, {\modelname} achieves $1.6\%$ better CM BLEU, $57\%$ better perplexity and $32\%$ better semantic coherence than the non-persona-based counterparts. The source code is available at: \url{https://github.com/victor7246/PARADOX}.
\end{abstract}

\section{Introduction}
\label{sec:intro}
\text{Code-mixing} ({\em aka} \text{code-switching}) appears when two or more languages are used interchangeably in a single utterance. It is common in multilingual societies like India, where more than $24\%$ of the population speaks in more than one language~\citep{sengupta_hit_2021}. Code-mixing is even more prevalent on social media. Informal usage of code-mixed languages on social media platforms like Twitter, Facebook, YouTube, and other online social networks gives rise to \textit{script-mixing}, in which a user can use a single script (\textit{e.g.,} Roman) or multiple scripts (\textit{e.g.,} Devanagari for Hindi and Roman for English) within the same text~\citep{srivastava_understanding_nodate}. Recent literature has made significant efforts to understand syntactic structure and semantics from code-mixed texts ~\citep{singh_language_2018, singh_twitter_2018, sengupta_comprehensive_2022}. Similar attempts have been made for pragmatic tasks -- humour, sarcasm and hate detection in the code-mixed regime~\citep{sengupta_does_2022, bansal-etal-2020-code}.

Text generation models need to understand the syntax and semantics of texts and preserve semantic coherence during generation. Previous studies utilized recurrent neural networks with generative models~\citep{zhang_adversarial_2017}, as well as self-attention-based pre-trained language models~\citep{zhang_dialogpt_2020} for generating monolingual texts. However, such an effort is limited in case of code-mixing. Previously, linguistic theories~\citep{pratapa_language_2018, gupta_semi-supervised_2020}, transfer learning~\citep{gupta_semi-supervised_2020}, and autoencoding~\citep{samanta_deep_2019} based approaches have been used to generate code-mixed texts from parallel corpora or reference data. 
However, none of these methods incorporates user information while generating code-mixed texts. Unlike traditional languages, code-mixing is a derived language whose adoption depends on different socioeconomic, demographic, and linguistic factors~\citep{rudra_understanding_2016, parshad_what_2016}. Figure~\ref{fig:user_data} demonstrates the code-mixing behaviour among Indian users on Twitter and  YouTube in terms of adoption and patterns in code-mixing. We visualize the mean and standard deviation of the Code-Mixing Index (CMI)~\citep{gamback_comparing_nodate} and the length of tweets/comments posted by different users. The distributions show how different users conceive and prefer code-mixing. 

User persona plays a vital role in generation models, particularly in personalized generation settings, such as conversational agents and recommendation engines. Several studies have contributed towards persona-based dialogue generation~\citep{zheng_pre-training_2019, wang_speaker-aware_2021}, personalized story generation~\citep{chandu_my_2019}, and other sub-tasks in text generation. Being a conversational language, personification of code-mixing could be deemed appropriate for conversational systems such as recommender engines, mental health counselling bots, and event booking applications.

\begingroup
\setlength{\intextsep}{0pt}%
\setlength{\columnsep}{5pt}%
\begin{wrapfigure}{r}{0.5\textwidth}
\centering
\includegraphics[scale=0.18]{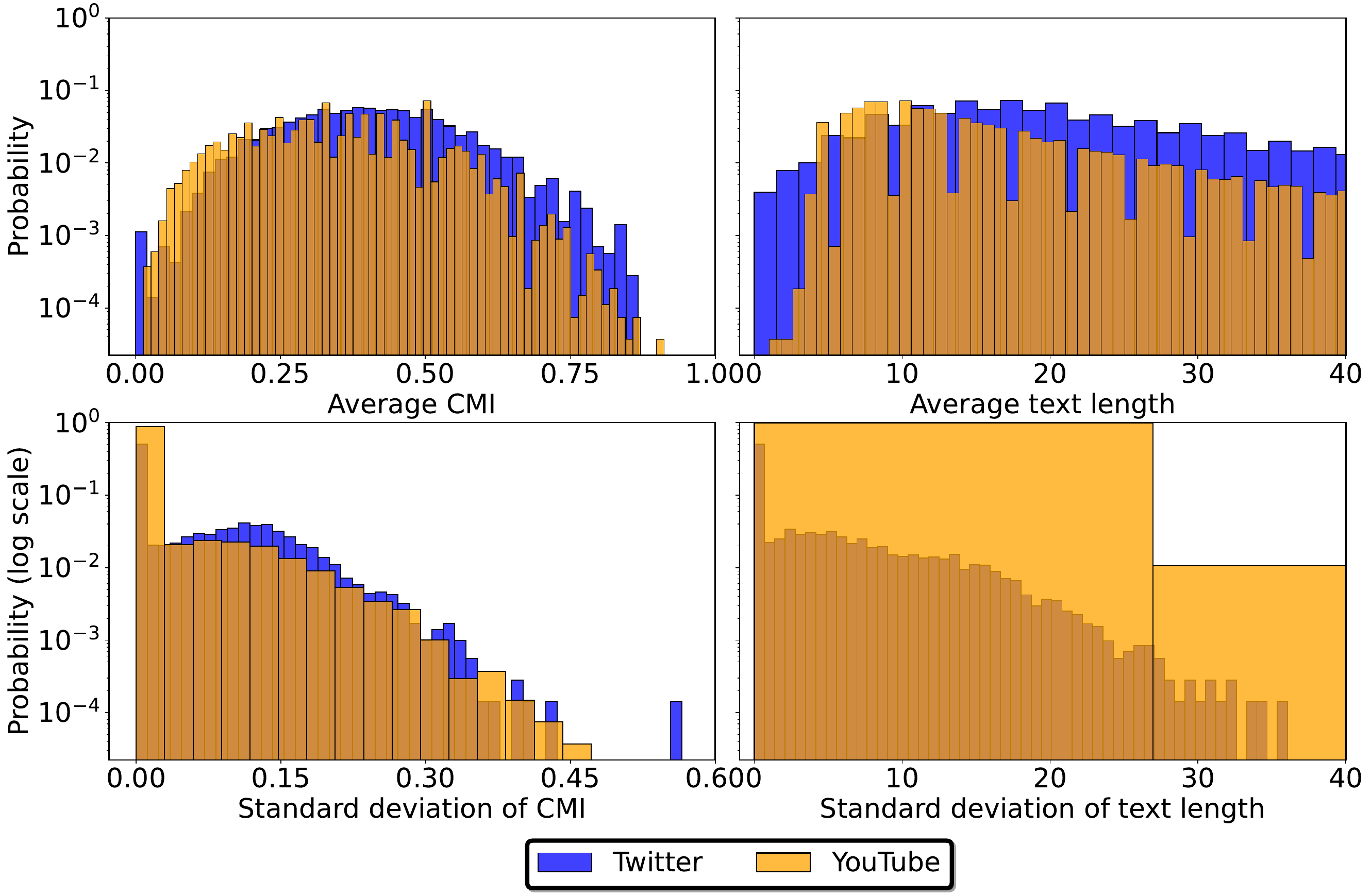}
\caption{User-specific distribution of Code-Mixing Index (CMI) and text lengths across different platforms. CMI is calculated as the fraction of minority language words in a text. For instance, the CMI of the text ``\color{black}{I don't want your} \color{black}{nautanki}'' (``I don't want your gimmick'') is $\frac{1}{5} = 0.2$, the fraction of Hindi (minority language in this example) words in the text. Texts skewed toward monolingualism, \textit{i.e.,} having an unequal proportion of words between different languages, tend to have lower CMI than multilingual texts.}
\label{fig:user_data}
\end{wrapfigure}

This motivates us to develop {\modelname}, a novel {\bf p}ersona-{\bf a}ware gene{\bf ra}tive mo{\bf d}el for c{\bf o}de-mi{\bf x}ed text generation.
It aims to generate personalized code-mixed texts by leveraging users' historical utterances. It uses a Transformer-based encoder-decoder architecture to learn the semantics of code-mixed generation. The model utilizes a novel persona encoder to encode a user persona from their behavioural preferences. Instead of projecting the user's persona onto a static space, {\modelname} projects it onto a probabilistic latent space and captures the contextual persona based on their historical persona. Additionally, \modelname\ uses an alignment module to re-align decoder outputs to generate coherent texts. 

We evaluate {\modelname} against the vanilla Transformer in terms of both the quality and coherence of generated texts. To quantify the extent of personification in the code-mixed generation, we propose four metrics -- CM BLEU, CM Rouge-1, CM Rouge-L, and CM KS (here {CM} stands for code-mixing). On average, {\modelname} achieves $1.6\%$ better CM BLEU than the non-persona counterpart. We also conduct a detailed human evaluation, concluding that {\modelname}-generated code-mixed texts are $32\%$ more semantically coherent than that of the vanilla Transformer model. \modelname\ can imitate a user's linguistic preference $4\%$ better than the non-persona-based Transformer model. Our empirical analyses also highlight the effectiveness of {\modelname} over pre-trained large language models. On average, {\modelname} achieves $4.2\%$ better CM BLEU, $11\%$ better CM Rouge-1, and $9.6\%$ better CM Rouge-L than the pre-trained Llama 2~\citep{touvron2023llama} and GPT-4~\citep{achiam2023gpt} models.

\noindent \textbf{Contributions.} The major contributions of this paper are summarized below:

\begin{itemize}[leftmargin=*,topsep=0pt,itemsep=-1ex,partopsep=1ex,parsep=1ex]

\item We make a pioneering effort in utilizing user persona in code-mixed text generation. Compared to existing approaches, {\modelname} does not require parallel corpora or reference data for code-mixed text generation.
\item We propose a probabilistic persona encoder module that learns the latent persona of a user from historical contexts. {\modelname} captures the user persona implicitly and does not require explicit persona features such as user demographic information to encode user behaviours.
\item We design an alignment module to automatically induce alignments between different subwords. Empirical results show that it improves the coherence of generated texts.
\item We propose three metrics influenced by supervised machine translation -- CM BLEU, CM Rouge-1, and CM Rouge-L for evaluating the personification of code-mixed generation models. We also propose CM KS as a distance measure to evaluate code-mixing generation models.
\item Finally, we collect two large-scale longitudinal datasets from Twitter and YouTube, primarily monolingual Hindi and Hindi-English code-mixed texts. The datasets will be valuable for code-mixing research.
\end{itemize}


\section{Related Works}
\label{sec:related_works}
\paragraph{\bf Rule-based and Linguistic Approaches.} Code-mixed text generation has garnered much interest in recent times. \citet{pratapa_language_2018} explored \textit{equivalence constraint} (EC) theory to generate Spanish-English code-mixed texts from monolingual corpora. Their linguistic theory-based approach showed superiority over recurrent neural networks in complex code-mixed text generation. \citet{rizvi_gcm_2021} developed  {GCM}, a toolkit that utilizes different linguistic theories to generate code-mixed texts. Motivated by embedding matrix theory, \citet{srivastava_hinge_2021} proposed rule-based methods to generate Hindi-English code-mixed texts. \citet{santy2021bertologicomix} utilized parse tree structures within the monolingual texts for generating code-mixed texts. Alternative approaches use generative models -- generative adversarial networks~\citep{goodfellow_generative_2014}, or variational autoencoder (VAE)~\citep{kingma_auto-encoding_2014}. Towards this, ~\citet{garg_code-switched_2018} explored recurrent neural networks with SeqGAN pre-training for generating Mandarin-English code-mixed data. \citet{samanta_deep_2019} developed a VAE-based method to generate realistic and coherent Hindi-English code-mixed texts. Other classes of code-mixed text generation models explore alignment within parallel corpora for code-mixed generation. Notably, ~\citet{winata2019code, tan2021code} explored word alignments and candidate selection from parallel corpora for generating synthetic code-mixed texts. \citet{amin2023marathi} explored word alignments for generating Marathi-English code-mixed text generation. On a similar attempt, ~\citet{dowlagar2021gated} explored gated convolutional encoder-decoder models to identify the compositional structure and translate English texts to Hinglish.

\paragraph{\bf Pre-trained Models.} With the inception of self-attention~\citep{vaswani_attention_2017}, several attempts have been made to develop large pre-trained models showing exceptional performances in semantic and generative tasks. Among these methods, multilingual models, such as XLM~\citep{lample_cross-lingual_2019}, XLM-RoBERTa~\citep{conneau_unsupervised_2020}, and mBART~\citep{liu_multilingual_2020} have shown noticeable performance even on low-resource languages. Recently, MuRIL~\citep{khanuja_muril_2021} was proposed, superseding the performances of multilingual-BERT~\citep{devlin_bert_2019} on different syntactic and semantic tasks on a diverse set of low-resource languages. \citet{gupta_semi-supervised_2020}   devised a semi-supervised approach to transfer knowledge from XLM to generate synthetic Hindi-English code-mixed texts. \citet{gautam2021comet} explored a pre-trained mBART model for generating Hindi-English code-mixed texts. \citet{jawahar2021exploring} explored multilingual text-to-text models with curriculum learning for generating Hindi-English code-mixed texts. They pre-trained an encoder-decoder model on synthetic code-mixed texts, which improved the generation quality on the gold code-mixed dataset. In a recent study, ~\citet{yong2023prompting} explored multilingual large language models (LLMs) for generating code-mixed texts in a zero-shot setting. They explored InstructGPT, ChatGPT~\citep{ouyang2022training}, BLOOMZ~\citep{muennighoff2022crosslingual} and Flan-T5-XXL~\citep{chung2022scaling} for generating code-mixed texts in Indonesian, Malay, Chinese, Tagalog, Vietnamese, Tamil, and Singlish. They further emphasized the importance of better prompt templates and language pairing for generating more coherent and natural code-mixed texts.

\color{black}

\paragraph{\bf Personification of Code-mixing Languages.} Language, as a mode of communication, is often personalized for different users~\citep{king-cook-2020-evaluating}. The personification of language arises naturally, depending on the context, socio-economic and demographic background of the user and their audience~\citep{reichelt2014talk}. Particularly in multilingual societies, the complex dynamics between different languages play a crucial role in the personification of language. ~\citet{sengupta2024social} highlighted the personalization aspects of Hindi-English code-mixed language, highlighting the importance of different sociological aspects behind the evolution of personalized code-mixed languages.

\color{black}

\paragraph{\bf Major Limitations of Existing Studies.} Despite their popularity in several generative applications, personalization remains neglected in the code-mixed generation. The existing code-mixed generation models utilize parallel corpora to understand the switching patterns and generate code-mixed texts synthetically. These limitations motivate us to develop a code-mixed language generation model that can automatically learn the language's semantics and capture the linguistic preferences of users while generating texts. Our proposed method, {\modelname}, improves the quality of generated texts and preserves the real-life phenomenon of code-mixing among users. 

\begin{figure*}
\centering
\includegraphics[scale=0.7]{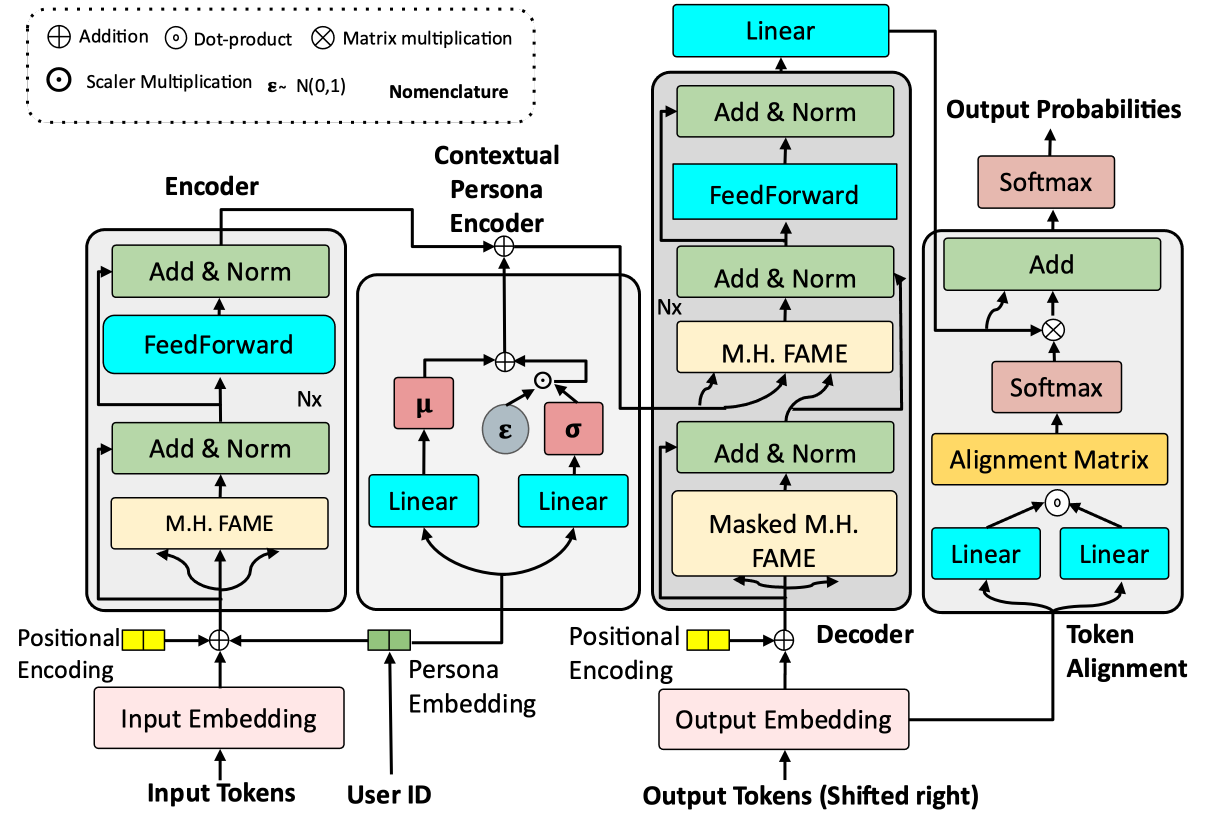}
\caption{{\modelname}: Transformer encoder-decoder architecture with persona encoder (multi-headed (M.H.)
fused attention (FAME)).}
\label{fig:model_architecture}
\end{figure*}

\section{Proposed Methodology}
\label{sec:method}

Here, we explain \modelname, and the personalized code-mixing (\text{CM}, henceforth) generation process utilizing user persona. Before describing the proposed generative model, we first elaborate on the definition of personalization of code-mixed language.

\color{black}
\subsection{Personalized Code-mixing}
As highlighted in Section~\ref{sec:related_works}, the study of the personalization of a language is crucial to develop conversational generative models. Typically, a personalized generative model captures the user's preferential behaviors during generation. Instead, our work delves into the personalization of user's linguistic preferences in terms of mixing multiple languages in an utterance. We define the personalization of code mixing as a behavioral language that captures a user's writing style given the historical trends, social context, targeted demographics, and topical relevance. The proposed method, \modelname, aims to capture these behavioral and contextual characteristics of user historical utterances and generate texts that are aligned with their linguistic preferences.

\color{black}
\subsection{{\modelname} Architecture}
{\modelname}, as shown in Figure~\ref{fig:model_architecture}, consists of four components -- (a) an encoder, (b) a contextual persona encoder, (c) a decoder, and (d) an alignment module. \modelname\ utilizes a persona encoder to implicitly encode a user's persona based on his/her historical utterances. It further projects the persona onto a probabilistic latent space to capture the user's contextual persona. Finally, {\modelname} employs an alignment module to re-calibrate the output sequences that help our model understand the language of code-mixing and enable the model to generate coherent texts. 

\subsubsection{The Encoder with FAME}
The encoder is used to jointly learn the semantics of a code-mixed text and a user's global persona based on the previous comments/tweets. A comment/tweet ${\mathbf X}_u$ by user $u$ is first tokenized using byte-pair encodings~\citep{sennrich_neural_2016} into $\langle x_{1}, x_{2}, ..., x_{n} \rangle$. The initial contextual embedding of a token $x_{i}$ is conditioned with the user persona as 

\begin{equation}
\widetilde{Emb_{(x_i,u)}} = Emb_{x_{i}} + PE_{i} + Emb_{u}
\end{equation} 
where $Emb_{x_{i}}$ is the initial token embedding, $PE_{i}$ is the positional encoding at position $i$, and $Emb_{u}$ is the user's global persona embedding captured through an embedding layer. \color{black} $Emb_{u}$ is computed with a unique identifier for each user. This unique user ID is important to encode the temporal evolution of a user persona. In order to calculate the global persona for cold-start users (users present in the development set but not during model training), we use a learnable \texttt{[UNK]} token embedding as $Emb_{u}$. \color{black} We use a stacked encoder, in which each encoder consists of {multi-headed fused attention} (FAME), followed by a {residual}, {\color{black}{layer}\color{black}-normalization}, and {pointwise feed-forward} layers. ~\cite{sengupta_hit_2021} introduced FAME by combining scaled dot-product attention~\citep{vaswani_attention_2017} and outer-product attention~\citep{le_self-attentive_2020} and showed to be effective in capturing both semantics and morphology of code-mixed texts. 

\subsubsection{Contextual Persona Encoder}

We project the static persona embedding onto a probabilistic latent space for generating the contextual persona embedding for each user in a given context. \color{black}{We hypothesize that each user has a static (global) persona and a contextual (local) persona. The motivation behind projecting the persona embedding to a latent space is to capture the contextual perturbations in the user persona. 
\color{black} For instance, Table~\ref{tab:user_behavior} highlights a user who predominantly uses monolingual English for raising political opinions (CMI $0.0$), however, at occasions switches between English and Hindi to create stronger narrative. 

\begin{table*}[ht!]
\centering
\vspace{2mm}
\scalebox{0.85}{
\begin{tabular}{c| p{30em}| c}
\hline
User ID & Generated Text & CMI\\
\hline
\multirow{6}{*}{158} & \textbf{CM:} This school student  is a passionate supporter of clean politics. & 0.0\\
& \textbf{Eng:} this school student  is a passionate supporter of clean politics. & \\
\cdashline{2-3}
& \textbf{CM:} once again vendetta politics will fail. \color{blue} Satyamev jayate. & 0.21\\
& \textbf{Eng:} once again vendetta politics will fail. Truth will triumph. & \\
\cdashline{2-3}
& \textbf{CM:} This so called party with difference has links with \color{blue} brastacharis. & 0.12\\
& \textbf{Eng:} This so called party with difference has links with corrupts. & \\
\hline
\end{tabular}}
\caption{Examples of user linguistic behavioral changes in different contexts. Usage of Hindi words are highlighted with \color{blue}{blue}. \color{black}The user predominantly uses monolingual English for expressing political opinions. However, at different instances the user uses Hindi interchangeably for referring to contextual popular narratives for strengthening their expression. } \label{tab:user_behavior}
\vspace{2mm}
\end{table*}

\color{black}Formally, we generate a contextual persona embedding 
\begin{equation}
\widetilde{Emb_{u}} \sim q_{\phi}(z|Emb_{u}) = \mathcal{N}(\mu_{u},\,\sigma_{u}^{2}). 
\end{equation} 
Towards this, we define two linear projection matrices to learn the distribution location and scale parameters as 
\begin{equation*}
\mu_{u} = Emb_{u} . W_{\mu} \text{  and  } \sigma_{u} = Emb_{u} . W_{\sigma}.
\end{equation*} 
Following the reparameterization trick~\citep{kingma_auto-encoding_2014}, we define the final generated persona encoding as:
\begin{equation}
\widetilde{Emb_{u}} = \mu_{u} + \epsilon_{u} \odot \sigma_{u}
\end{equation} 
where $\epsilon_{u}$ is the random noise, independently drawn from $\mathcal{N}(0,1)$. \color{black} We refer to this method \textbf{randomized persona encoder}. In another ablation, we use only the linear projection $\widetilde{Emb_{u}} = \mu_{u} = Emb_{u} . W_{\mu}$ for representing the contextual persona embedding, which we call \textbf{linear persona encoder}. \color{black} We obtain the hidden representation of token $x_{i}$ conditioned on the contextual persona encoding as 
\begin{equation}
\widetilde{h_{(x_{i},u)}} = h_{(x_{i},u)} + \widetilde{Emb_{u}}
\end{equation} 
where $h_{(x_{i},u)}$ is the final hidden representation obtained from the final layer of the encoder.

\subsubsection{The Decoder}
We adopt the Transformer decoder conditioned on the contextual user persona. Similar to the original Transformer decoder, we use a stacked decoder initialized with the encoded output sequence. Drawing the motivation from autoregressive generative language models like GPT2~\citep{radford_language_nodate}, the decoder's objective is to predict the next token, conditioned on all the previous tokens. The input to the decoder is the encoded input sequence \color{black}{added} \color{black} with positional encoding. Each decoder block consists of masked multi-headed FAME, a residual connection, and a normalization layer. We also deploy multi-headed FAME to attend to each decoder token with the encoded input tokens $\widetilde{h_{(x_{i},u)}}$. For each decoder input position $j$, we generate a hidden representation $h_{(j,u)}^{(dec)} \in {\mathbb{R}}^{|V|}$, representing the output token at $(j+1)^{th}$ position; $|V|$ is the vocabulary size of the decoder.

\subsubsection{The Alignment Module}
The final layer of {\modelname} is an alignment module that learns the latent alignment matrix and re-aligns the outputs generated by the decoder. The primary objective behind using alignments in generative models is explicitly learning the global semantic similarity between different tokens. We use two projection matrices, $W^{Q}$ and $W^{K}$, to project the decoder token embedding matrix $Emb^{(dec)}$ into two different subspaces. The alignment matrix is defined as,
\begin{equation}\small
    \mathcal{A} = softmax \left(\frac{Q \cdot K^{T}}{\sqrt{d}} \right)
\end{equation}
where $Q = Emb^{(dec)} \cdot W^{Q}$,     $K = Emb^{(dec)} \cdot W^{K}$, and $d$ is the hidden size of the decoder. This operation resembles the scaled dot-product attention mechanism~\citep{vaswani_attention_2017}. However, as opposed to attention, we compute the global context by considering the original embedding space of all tokens. Finally, the re-aligned hidden representation is derived as,
\begin{equation}
     \widetilde{h_{(j,u)}}^{(dec)} = h_{(j,u)}^{(dec)} \cdot \mathcal{A} + h_{(j,u)}^{(dec)}
\end{equation}
This hidden representation is finally fed to a softmax layer to convert the outputs into probabilities. \color{black}{It is important to notice that the alignment matrix $\mathcal{A}$ is not intended to capture the semantic similarity between contextual tokens but rather to learn their similarities at a global level. By doing so, the alignment module can capture the associations between all the tokens and recalibrate their probabilities during decoding. Moreover, as the dot-product is computed over the embedding matrix, the autoregressive rule of text generation is not violated here.}

\color{black}
For the sake of simplicity, we denote the combination of text encoder and persona encoder as `encoder' and the combination of Transformer decoder and the alignment module as `decoder' throughout this paper. The generative model is trained w.r.t. the decoder reconstruction cross-entropy loss. The contextual persona encoder gives rise to a variational KL-divergence loss. We use a variational hyperparameter $\lambda$ to assign its weightage in the final computed loss.

\subsection{Training Curricula}

To learn the model parameters, we primarily minimize the reconstruction loss on output sequence $\langle y_{1}, y_{2}, ..., y_{m} \rangle$ between defined as,
\begin{equation}\small
{\mathcal{L}_1}^{(x,u)} = \sum_{j=1}^{m}{y_{(j,u)} \log(P_{\theta_{dec}}(y_{(j,u)}|{\mathbf Y}_{(0:j-1,u)},{\mathbf X}_{1:n}, u))}    
\end{equation}
The output sequence is initialized with $y_0 =$ \texttt{[CLS]} token. The contextual persona encoder module arises a Kullback–Leibler divergence loss between the variational distribution and true posterior distribution, which can be derived~\citep{kingma_auto-encoding_2014} to
\begin{equation}\small
{\mathcal{L}_2}^{(u)} = 
-\frac{1}{2}  \sum_{k=1}^{d} \left (1 + 2 \cdot \log({\sigma^{k}_{u}}) - ({\mu_{u}^{k}})^{2} - ({\sigma_{u}^{k}})^{2} \right)
\end{equation}
During training, we minimize the task-specific loss
\begin{equation}\small
    \mathcal{L}^{(x,u)} = {\mathcal{L}_1}^{(x,u)}  + \lambda \cdot {\mathcal{L}_2}^{(u)}
\label{eq:main}
\end{equation}
for each text and user id pair $(x,u) \sim \mathcal{D}$ on training data. The persona encoding weight $\lambda$ is a hyperparameter we set before running the experiments. 


\SetKwComment{Comment}{/* }{ */}
\SetKw{Required}{Require:}
\SetKw{Return}{Return}

\begin{algorithm}[!t]
\SetAlgoLined
\caption{Code-Mixed Text Generation with {\modelname}}\label{algo:generation}
\Required{Trained model $\mathcal{M} = (enc, dec)$, user id $u$, historical utterance $x_u$, prompt word $\{w_1\}$, decoder vocabulary $V$}

\Required{$max\_length \in \mathbb{N}$}

$L \leftarrow \{\texttt{[CLS]}, w_1\}$;

$\Tilde{w} \leftarrow \emptyset$;

$i = m$;

\While{$\Tilde{w} \neq$ \texttt{[SEP]} and $i < max\_length$}{
  $h_{(x_{u},u)} = enc(x_{u},u)$;
  
  $P_{i+1} = dec(L,h_{(x_{u},u)})$;
  
  $\Tilde{w} \leftarrow \operatorname*{arg\,max}_V P_{i+1}$;
  
  $L \leftarrow L \cup \{\Tilde{w}\}$;
}

\Return{L}
\end{algorithm}

\subsection{Code-Mixed Generation}
We adopt an autoregressive generation technique to generate new code-mixed texts for different users. To encode the user's historical persona, we use the user's last utterance (comment/tweet) in the encoder. 
\color{black} As \modelname\ is trained autoregressively on all historical utterances for different users, it can capture the entire conversational history of a user from the last utterance without passing the entire history during generation. A typical text generation model starts decoding with a placeholder \texttt{[CLS]} token. Instead, we pass an additional seed word as input to the decoder for a more guided generation. \color{black} We formally report the text generation process in Algorithm~\ref{algo:generation}.

\section{Experimental Setup}
This section elaborates on the experimental setup we adopt to evaluate our model and baselines on personalized code-mixed generation.

\subsection{Datasets}
To the best of our knowledge, no existing longitudinal dataset is available for Hindi-English code-mixed. A longitudinal dataset is required to study the temporal evolution of a language. 
Although some datasets in the literature consist of Hindi-English code-mixed texts collected from various online social networks, none of them contain user-specific information, making them unsuitable for our study. To overcome this, we collected code-mixed texts from the two most popular mediums where Indians are engaged -- Twitter and YouTube. From Twitter, we collected over $0.8$ million in tweets starting from the year $2011$ till date, from which we filtered only tweets originating from Mumbai and Delhi metropolitan regions, two cities with the largest Hindi population. We used Twitter API for academic research with full archival access\footnote{\url{https://api.twitter.com/2/tweets/search/all}}. Further, for relevance, we restricted ourselves to tweets related to `Cricket', `Bollywood', `Politics', and `Government'. Starting at $2014$, Twitter automatically tags the language of a tweet. We selected tweets with only non-empty language tags. This gives us a total of $226,480$ tweets from $19,782$ users.

From YouTube, we chose two channels -- \textit{NishaMadhulika}\footnote{\url{https://www.youtube.com/c/nishamadhulika}} (a popular chef based out of India with more than $12.7$ million followers), and \textit{T-Series}\footnote{\url{https://www.youtube.com/aashiqui2}} (a popular Hindi music record channel started in $1983$ having more than $200$ million followers). We selected $42$ videos from the NishaMadhulika channel and $69$ from the T-Series that were first posted in $2011$. We scraped all comments corresponding to these videos, accounting for $144,822$ comments from $99,998$ users.

For both datasets, we use a pre-trained language model open-sourced with Huggingface\footnote{\url{https://huggingface.co/sagorsarker/codeswitch-hineng-lid-lince}}, that was fine-tuned on Hindi-English parts-of-speech (PoS) and language identification (LID) tasks. Using this model, we label each token in each text with the corresponding language (Hindi or English) and their associated PoS. We tag a text as code-mixed only when the text contains at least one Hindi verb written in either Devanagari or Roman script. We select users who have at least three utterances in their entire timeline. Finally, we are left with $18,126$ tweets (from $2,241$ users) and $8,957$ YouTube comments (from $1,349$ users).

We remove all the HTML tags, URLs, emoticons, user mentions (starting with `@'), and hashtags (starting with `\#'). For simplicity, we remove all numeric values from texts, as well. Finally, we convert all texts to lowercase. We highlight the key statistics of the datasets in Table~\ref{tab:dataset}. We use a $75$-$25$ split for training and validation \color{black}{with stratified sampling. Therefore, we can ensure at least one training and validation sample for each user. We choose the first word for each validation sample as the seed word for generating the code-mixed text. Finally, the generated code-mixed text is evaluated against the original validation text using the proposed personalization evaluation metrics.} 

\color{black}

\begin{wraptable}{r}{0.5\textwidth}
\resizebox{0.5\columnwidth}{!}{
\begin{tabular}{l c c c c}
\hline
\textbf{Dataset} & \textbf{\#Texts} & \textbf{\#Users} & \textbf{Mean text length} & \textbf{Mean CMI} \\\midrule
Twitter & 18126 & 2241 &\multicolumn{1}{c}{21.77} & 0.41 \\
YouTube & 8957 & 1349 &\multicolumn{1}{c}{28.89} &0.36 \\
\hline
\end{tabular}}
\caption{Dataset statistics, with mean text CMI and the average text lengths, demonstrating the extent of code-mixing.} \label{tab:dataset}
\end{wraptable}

\subsection{Evaluation Metrics}\label{sec:eval}
We adopt intrinsic and extrinsic evaluation metrics to evaluate our model in terms of semantic understanding of code-mixing language and the ability to personify code-mixing for different users. 

For the intrinsic evaluation, we use {\bf perplexity}, a metric that measures the predictive power of a language model, compared against ground truth. We calculate perplexity as $e^{loss}$, with $loss$ being the cross-entropy reconstruction loss on the validation data. A lower perplexity score indicates better reconstructibility and ability to learn the semantics of a generative model. 

{\color{black} Unlike~\cite{gupta_semi-supervised_2020}, we do not have any labelled gold data for evaluating our generative model. Therefore, traditional supervised evaluation metrics -- BLEU~\citep{papineni2002bleu}, Rouge~\citep{lin2004rouge} can not be used directly to evaluate the personification aspects of code-mixed generation models. Similarly, other extrinsic evaluation measures such as Multilingual index (M Index)~\citep{barnett2000}, Burstiness and Span Entropy~\citep{guzman_metrics_2017} can not be used, as these metrics are predominantly used to evaluate the ability to capture corpus-level switching patterns of generative models. To overcome the limitations of the existing evaluation metrics, we propose four metrics for benchmarking generated code-mixed texts against the historical utterances by different users. \color{black} These proposed metrics calculate the similarity between the linguistic patterns of model-generated texts and the user's historical utterances. \color{black} We devise \textbf{CM BLEU} by calculating the BLEU score between the candidate and reference language sequences. For example - consider a candidate code-mixed text ``\it{mujhe} \color{black}{park} \color{black}{janaa} hai''} (``I want to go to the park'') and a reference text ``\textit{mujhe} \color{black}{rice} \color{black}{\textit{aur}} \color{black}{curry} \color{black}{khana hai}'' (``I want to eat rice and curry''). Using the LID model, we can extract the corresponding language sequences \{Hi, {Hi, Hi, Hi\} and \{Hi, En, Hi, En, Hi, Hi, Hi, Hi, Hi, Hi, Hi \} from the candidate and reference texts, respectively (here, Hi and En stand for Hindi and English, respectively). Therefore, considering only the unigram and bigram overlaps between the candidate and the reference, we calculate the CM BLEU score\footnote{Can be calculated and validated using \url{https://www.nltk.org/api/nltk.translate.bleu_score.html}} of $0.606$. If we use a different reference text ``\color{black}{I don't want your} \color{black}{nautanki}'' (Translation - ``I don't want your gimmick'') with the corresponding language sequence \{En, En, En, En, Hi\}, the CM BLEU reduces to $0.218$. The proposed metric could calculate the similarity between the switching patterns demonstrated in the candidate text and the historical references by calculating the overlap between the language sequences. Similarly, we compute \textbf{CM Rouge-1} and \textbf{CM Rouge-L} by computing Rouge-1 and Rouge-L scores between the candidate and reference language sequences. 

Additionally, we leverage the user-level historical CMI to evaluate the linguistic patterns of the generated texts. If a user historically prefers monolingualism over multilingualism, we want the generative model to learn the pattern and generate texts with a lower CMI value for the user. Towards this, we propose \textbf{CM KS}, a metric that computes the Kolmogorov-Smirnov distance between the generated and original CMI distribution of users. 
We highlight the relationships between these metrics by calculating the Pearson correlation between these measures, reported in Figure~\ref{fig:heatmap} of Appendix~\ref{sec:appx_eval_metric}. Strong negative correlations between perplexity and CM BLEU, CM Rouge-1, and CM Rouge-L indicate that understanding semantics is essential to personify and replicate the switching patterns. Therefore, by learning semantics well, the generative models can learn the code-mixing patterns for different users and generate texts that imitate users' linguistic patterns. On the other hand, the correlations between CM KS and other metrics are weak, indicating that the linguistic preferences of users have no apparent linear relationships with switching patterns. \color{black} These four metrics capture different aspects of the personalized usage of code-mixing for different users, as introduced in Section~\ref{sec:method}.

\vspace{-1mm}
\color{black}
\subsection{Baseline Methods}
\vspace{-1mm}

We consider several code-mixed generation models for comparative evaluation.

\noindent $\rhd$  \textbf{VACS}~\citep{samanta_deep_2019}  is a VAE-based encoder-decoder model, primarily developed for generating Hindi-English synthetic code-mixed texts.

\noindent $\rhd$  \textbf{GCM}~\citep{rizvi_gcm_2021}  toolkit uses several linguistic theories and heuristics to generate code-mixed texts.

\noindent $\rhd$  \textbf{CM-XLM}~\citep{gupta_semi-supervised_2020}  is a generative model that utilizes pre-trained multilingual language model XLM to generate Hindi-English code-mixed texts from parallel corpora.

These code-mixed generation models consider monolingual reference data for generating code-mixed texts and generate code-mixed texts at a corpus level. Therefore, we compare these baselines only in intrinsic evaluation. We also utilize several self-attention-based encoder-decoder and pre-trained language models to evaluate the personification aspects of generative models.

\noindent $\rhd$ \textbf{Transformer}~\citep{vaswani_attention_2017} is an encoder-decoder architecture utilizing self-attention mechanism that has shown superior performances in generation tasks like machine translation.

\noindent $\rhd$ \textbf{MuRIL}~\citep{khanuja_muril_2021} is an encoder-based language model pre-trained on 17 Indian languages with a masked language modeling objective.

\noindent $\rhd$ \textbf{BLOOMZ}~\citep{muennighoff2022crosslingual} is a family of large language model based on multilingual BLOOM~\citep{workshop2023bloom} that was fine-tuned with multitask prompting. We use the $3$B parameter BLOOMZ model as our baseline.

\noindent $\rhd$ \textbf{Llama 2}~\citep{touvron2023llama} is a family of autoregressive large language models trained with reinforcement learning with human feedback (RLHF). We adopt the $13$B parameter instruction-tuned model as one of our baselines.

\color{black}
\noindent $\rhd$ \textbf{GPT-4}~\citep{achiam2023gpt} is a large multimodal model, accepting image and textual data and generating text outputs through autoregressive generation.

\color{black}

These pre-trained language models are only utilized in the extrinsic evaluation. We provide the hyperparameter settings for \modelname\ and all the baseline methods in Appendix~\ref{sec:appx_training_details}. 



\section{Comparative Analysis} 
In this section, we report the performances of {\modelname} and the non-persona-based code-mixed generation models in terms of the intrinsic and extrinsic evaluation measures. 

\begin{wraptable}{r}{0.6\textwidth}
\centering
{\scalebox{1}{\begin{tabular}{lrr}
\toprule
\textbf{Model} &\multicolumn{2}{c}{\textbf{Perplexity} $\downarrow$}  \\\cmidrule{2-3}
\textbf{} &\textbf{Twitter} &\textbf{YouTube} \\\midrule
GCM* & 4331.85 &4323.15  \\
CM-XLM* & 5413.22 &1603.59 \\
VACS & 361.05 & 552.35 \\
\hline
Transformer & 680.07 & 473.84 \\
\hline
\rowcolor{maroon!10}  {\modelname} & \textbf{297.43} & \textbf{196.21}  \\
 \hdashline
\rowcolor{maroon!10}  (-) Contextual Persona & 320.79 & 295.26 \\
\rowcolor{maroon!10}  (-) Speaker ID & 582.24 & 371.70 \\
\rowcolor{maroon!10}  (-) Alignment & 377.63 & 294.83 \\
\rowcolor{maroon!10}  (-) FAME & 864.98 & 583.09 \\
\bottomrule
\end{tabular}
}}
\caption{Intrinsic evaluation of the competing models based on perplexity ($\downarrow$: lower value indicates better performance). For models highlighted with *, perplexity is calculated with word-level generation. \color{black} \textbf{Bold} indicates the best results among all the models.}
\label{tab:main_result_perplexity}
\end{wraptable}

We report the intrinsic evaluation results in Table~\ref{tab:main_result_perplexity}. {\modelname} achieves $56\%$ better perplexity on the Twitter dataset than the vanilla Transformer. On the YouTube dataset, the margin is even higher ($59\%$). A lower validation perplexity shows \modelname's strong ability to understand code-mixing semantics and generate texts of different linguistic variations. {\modelname} achieves $18\%$ better perplexity on the Twitter dataset than the best non-transformer baseline VACS. On the YouTube dataset, the margin is significantly higher ($64\%$). 

Table~\ref{tab:main_result} highlights the extrinsic measures across all the generative models. On the Twitter dataset, {\modelname} achieves $2.4\%$ better CM BLEU than the Transformer model. Similarly, on the YouTube dataset, {\modelname} performs the best among all the baselines and outperforms the Transformer model with a margin of $2.0\%$. In terms of the Rouge measures, {\modelname} performs consistently better than the non-persona counterpart with an average margin of $1.9\%$. In terms of distance-based measures, \modelname\ performs significantly better than the Transformer model on both datasets. Overall, {\modelname} achieves $11\%$ lower CM KS distance than Transformer. Lower KS distance indicates the importance of utilizing user persona in generating user-specific code-mixed texts.


Among the pre-trained language models, fine-tuned Llama 2 and GPT-4 are the most competitive. 
Interestingly, even with a single example in the prompt (1-shot), CM BLEU increases by $9.8\%$ for Llama 2. Similar performance improvements are also observed with other extrinsic metrics. However, both Transformer and {\modelname} perform significantly better than the pre-trained language models in terms of personalized code-mixed text generation. On average, {\modelname} achieves $7.6\%$ better CM BLEU and $12.5\%$ better CM Rouge-1 than Llama 2. \color{black}{PARADOX even achieves $3.9\%$ better CM BLEU than the fine-tuned Llama model. Similar performance improvements are observed with CM Rouge-1 and CM Rouge-L metrics. 
PARADOX achieves $4.1\%$ better CM Rouge-1 and $3.8\%$ better CM Rouge-L than the GPT-4 model. Even with CM KS, our model outperforms GPT-4 with a wide margin of $14\%$. \color{black} Interestingly, with few-shot in-context learning, GPT-4 model demonstrates stronger performance than the existing baselines. On Twitter dataset, 1-shot GPT-4 model achieves higher CM BLEU and CM Rouge scores than \modelname. This superior performance can be justified with the facts that tweets are generally shorter and more monolingual (lower CMI), therefore, capturing the corpus-level trends are easier for robust LLMs like GPT-4 with even a single example in the prompt. However, it is worth noting that 1-shot GPT-4 still underperforms than \modelname\ in terms of the personalization metric -- CM KS. On YouTube dataset however, \modelname\ achieves $0.8\%$ higher CM BLEU and $1.5\%$ higher CM Rouge scores than the 1-shot GPT-4 model demonstrating its superiority in understanding and generating personalized code-mixed texts.

\color{black}

\begin{table*}[!t]
\centering
\scalebox{0.85}{
{\begin{tabular}{lrrrrrrrr}
\toprule
\textbf{Model} &\multicolumn{2}{c}{\textbf{CM BLEU} $\uparrow$} &\multicolumn{2}{c}{\textbf{CM Rouge-1} $\uparrow$} &\multicolumn{2}{c}{\textbf{CM Rouge-L} $\uparrow$} &\multicolumn{2}{c}{\textbf{CM KS} $\downarrow$} \\\cmidrule{2-9}
\textbf{} &\textbf{Twitter} &\textbf{YouTube} &\textbf{Twitter} &\textbf{YouTube} &\textbf{Twitter} &\textbf{YouTube} &\textbf{Twitter} & \textbf{YouTube} \\\midrule
MuRIL & 9.92 & 9.85 & 26.93 & 21.10 & 23.65 & 19.63 & 0.42 & \textbf{0.23}\\
BLOOMZ & 14.20 & 23.87 & 49.22 & 56.61 & 45.93 & 55.09 & 0.40 & 0.30 \\
Llama 2 (zero-shot) & 19.97 & 7.25 & 48.91 & 30.86 & 43.74 & 28.72 & 0.56 & 0.43\\
Llama 2 (1-shot) & 26.69 & 20.03 & 55.17 & 46.08 & 49.57 & 43.17 & 0.50 & 0.39\\
\color{black} Llama 2 (fine-tuned) &  \color{black}  21.97 & \color{black}  26.09 & \color{black} 55.89 & \color{black} 58.24 & \color{black} 51.07 & \color{black} 55.17 & \color{black} 0.40 & \color{black} \textbf{0.34} \\
\color{black} GPT-4 (zero-shot) & \color{black} \color{black} 30.94 & \color{black} 30.33 & \color{black} 57.46 & \color{black} 57.88 & \color{black} 50.89 & \color{black} 53.69 & \color{black} 0.42 & \color{black} 0.39 \\
\color{black} GPT-4 (1-shot) & \color{black} \color{black} \textbf{31.90} & \color{black} 30.57 & \color{black} 60.67 & \color{black} 61.46 & \color{black} \textbf{54.55} & \color{black} 57.34 & \color{black} 0.42 & \color{black} 0.36 \\
\hline
\color{black} Transformer & 22.21 & 29.36 & 58.69 & 61.02 & 51.10 & 57.27 & 0.42 & 0.37 \\
\hline
\rowcolor{maroon!10}  {\modelname} & 24.58 & \bf 31.37 & 60.60 & \bf 62.90 & 53.03 & \bf 59.16 &  0.36 & \bf 0.34  \\
\hdashline
\rowcolor{maroon!10}  (-) Contextual Persona & 24.06 & 30.67 & 60.03 & 62.44 & 52.52 & 58.71 & 0.35 & \bf 0.34  \\
\rowcolor{maroon!10}  (-) Speaker ID & 20.10 & 31.01 & 56.46 & 62.72 & 49.71 & 58.96 & 0.35 & \bf 0.34\\
\rowcolor{maroon!10}  (-) Alignment & 24.37 & 31.08 & \bf 60.79 & 62.46 & 53.18 & 59.15 &  \bf 0.32 & 0.37\\
\rowcolor{maroon!10}  (-) FAME & 18.49 & 28.29 & 54.01 & 60.67 & 47.68 & 57.10 & 0.36 & 0.37 \\
\bottomrule
\end{tabular}
}}
\caption{Extrinsic evaluation of pre-trained language models, Transformer and {\modelname} in terms of preserving user-level switching patterns ($\uparrow$ ({\em resp.} $\downarrow$): higher ({\em resp.} lower) value indicates better performance). \color{black}\textbf{Bold} indicates the best results among all the models.}
\label{tab:main_result}
\vspace{-5mm}
\end{table*}

\color{black}

\begingroup
\setlength{\intextsep}{0pt}%
\setlength{\columnsep}{5pt}%
\begin{wrapfigure}{l}{0.6\textwidth}
\centering
\includegraphics[scale=0.45]{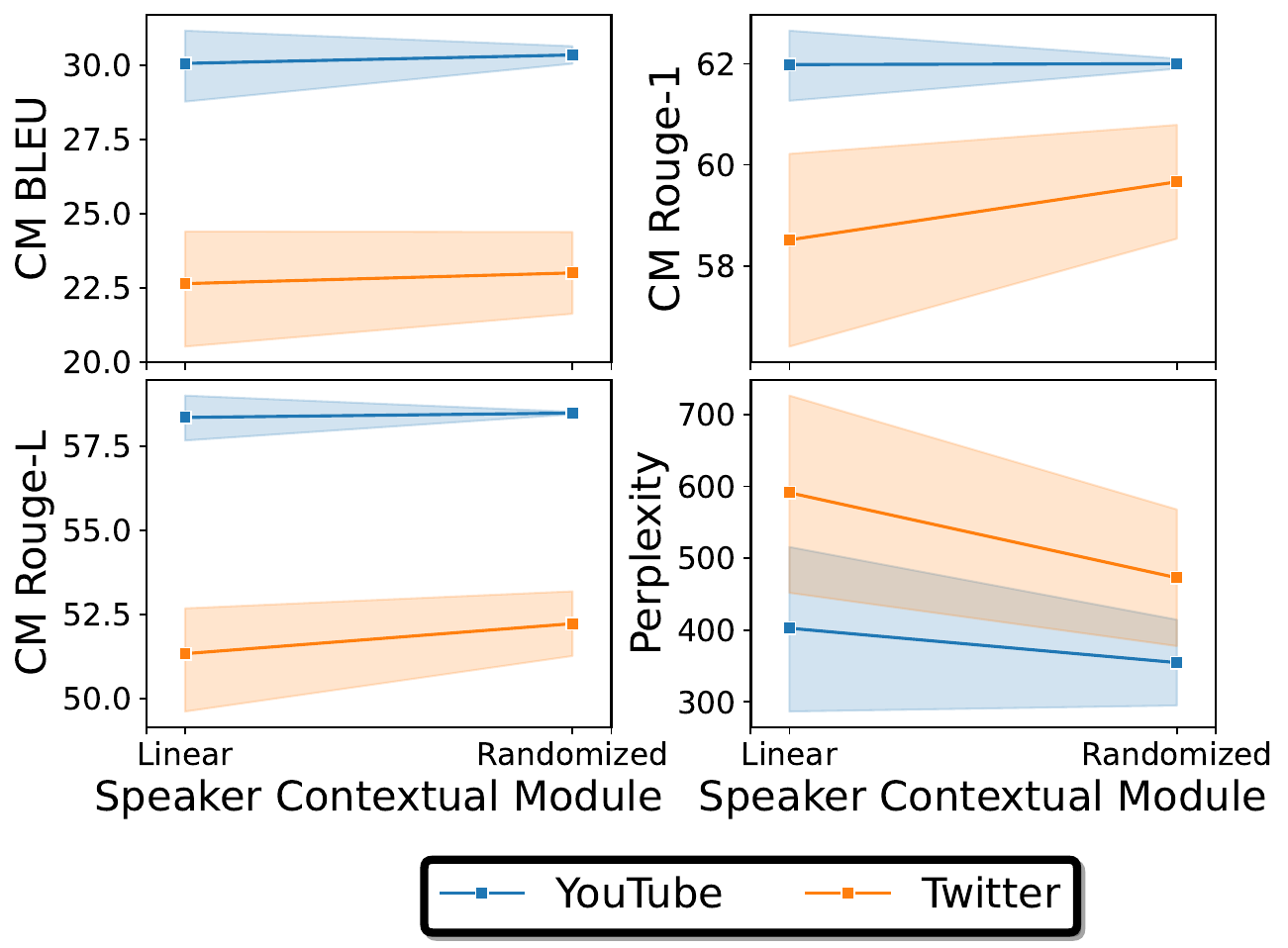}
\caption{Performances of \modelname\ under linear and randomized persona encoder.}
\label{fig:persona_module}
\end{wrapfigure}

Our ablation study in Table~\ref{tab:main_result_perplexity} shows the effectiveness of fused attention, contextual persona module, and the alignment module in {\modelname}. \color{black} Additionally, we empirically highlight the necessity of speaker ID encoding for personalized code-mixed text generation. \color{black} Adding FAME improves validation perplexity by $66\%$. Similarly, the contextual user persona and alignment modules improve validation perplexity by $21\%$ and $28\%$, respectively, justifying their contributions to modelling low-resource language. \color{black} Modeling the user persona with an additional identifier (speaker ID) improves the validation perplexity by $48\%$, highlighting the importance of the module for personalized text generation. 
Ablation results in Table~\ref{tab:main_result} also suggest the importance of different modules of \modelname\ for personalized code-mixed text generation. \color{black} Removing the contextual persona encoder, leads to an average performance drop by $0.52\%$ in regards to CM BLEU and CM Rouge scores. Moreover, an one-sided t-test concludes the statistical significance ($p$-value $< 0.001$) of the performance drop across different extrinsic measures and datasets. \color{black} Figure~\ref{fig:persona_module} highlights the distribution of intrinsic and extrinsic metrics for linear and randomized speaker contextualization modules. Not only does having a randomized speaker contextual module improve the performance, but it also improves the robustness of the generation model. Therefore, we argue that a random exploration of user persona helps our model in capturing the variability in linguistic patterns demonstrated by users in different contexts, enabling it to imitate the linguistic patterns during generation.


\color{black}
\subsection{Human Evaluation} 

\newcommand{\resulttab}{\begin{tabular}{lcc}\toprule
\textbf{Model} & \textbf{Semantic Coherence $\uparrow$} & \textbf{Linguistic Quality $\uparrow$} \\
\toprule
Transformer & 2.34 & 2.32 \\
\rowcolor{maroon!30} {\modelname} & \textbf{3.08} & \textbf{3.00} \\
\bottomrule
\label{tab:human_evaluation_results}
\end{tabular}}

\newcommand{\fleisstab}{\begin{tabular}{lcc}\toprule
\textbf{Model} & \textbf{Semantic Coherence} & \textbf{Linguistic Quality} \\
\toprule
Transformer & 0.14 & 0.12 \\
\rowcolor{maroon!30} {\modelname} & 0.11 & 0.15 \\
\bottomrule
\end{tabular}}

\begin{wraptable}{r}{0.55\textwidth}
\centering
  {\scalebox{0.8} \resulttab}%
  \vspace{-5mm}
  \caption{Human evaluation of the models.}
  \label{tab:human_evaluation}%
\end{wraptable}

We perform a human evaluation study to evaluate the code-mixed texts generated by {\modelname} and the vanilla Transformer. \color{black} In the previous section, we discussed our methodology for evaluating the personification capabilities of \modelname\ using the intrinsic and proposed extrinsic metrics. With human evaluation, we aim to assess the generative models in terms of linguistic correctness and contextual relevance. \color{black} Towards that, we randomly sample $24$ examples from each of these models and ask $30$ human evaluators\footnote{Evaluators are proficient with Hindi-English code-mixed language and their age ranges in $21-35$.} to rate these examples based on \textit{Semantic coherence} and \textit{Linguistic quality}. Semantic coherence measures the meaningfulness of the code-mixed texts, whereas, with linguistic quality, we measure their structural validity. Both the scores ranged between $1$-$5$, $1$ being the lowest, and $5$ being the highest. 

Table~\ref{tab:human_evaluation} presents the average semantic coherence and linguistic quality scores, along with Fleiss's Kappa~\citep{fleiss1971measuring} scores among the annotators. We observe that {\modelname} displays a better semantic coherence ($32\%$ better), as well as better linguistic quality ($29\%$ better) than the Transformer model. We observe fair agreement (Kappa $0.13$ for semantic coherence and Kappa $0.14$ for linguistic quality) among the annotators for both models. 

\section{Analysis of Code-Mixed Generation}

\begingroup
\setlength{\intextsep}{0pt}%
\setlength{\columnsep}{5pt}%
\vspace{-2mm}
\begin{wrapfigure}{r}{0.6\textwidth}
\centering
\includegraphics[scale=0.27]{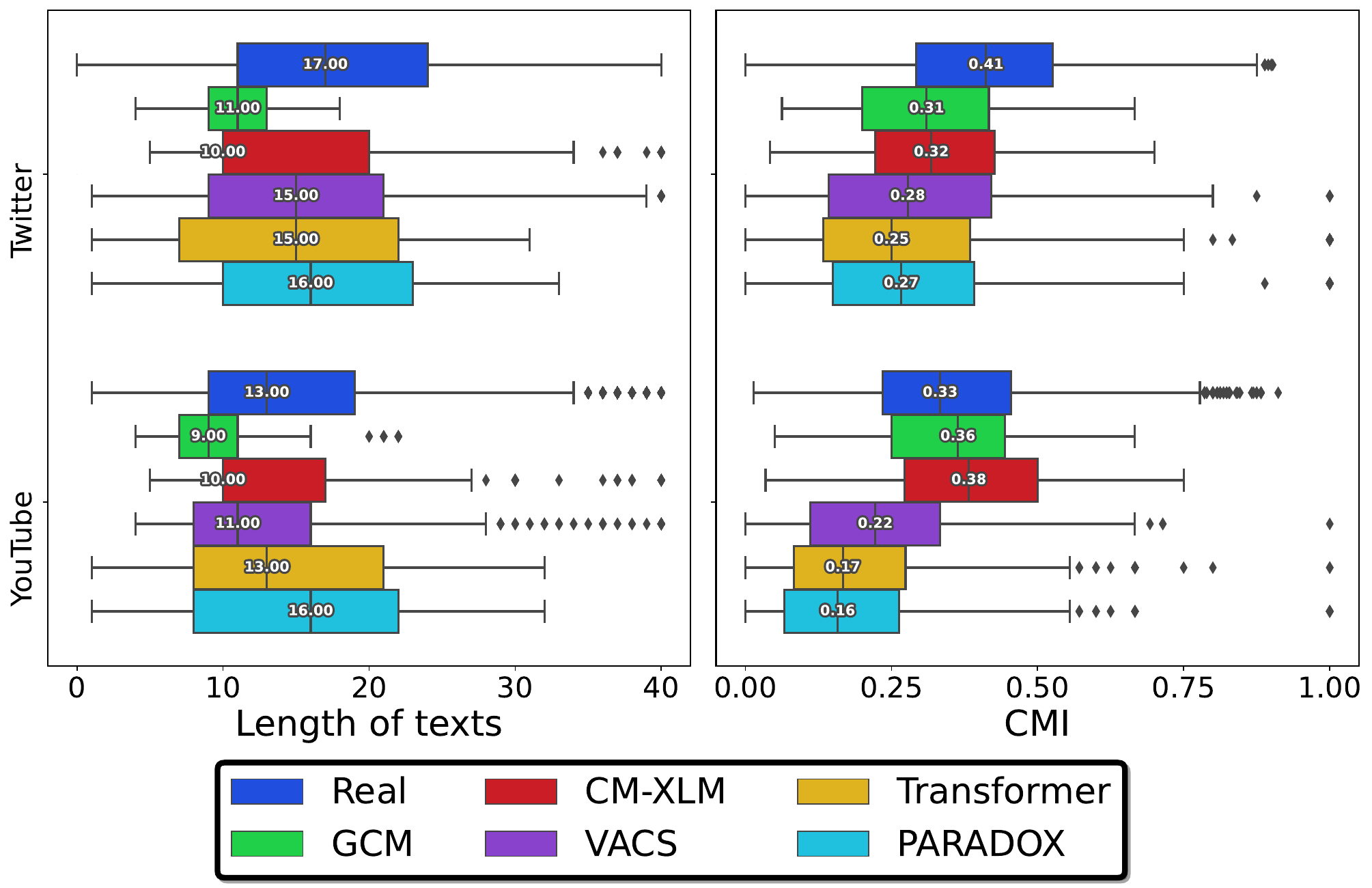}
\caption{Comparison of different text generation models.}
\label{fig:analysis}
\end{wrapfigure}

We further study the quality of code-mixed generation and compare them against other baselines. We analyze the distribution of length and CMI of texts generated by different generative models and report in Figure~\ref{fig:analysis}. Trained on the Twitter dataset, {\modelname} generates texts with a median length of $16$, $25\%$ higher than the other generative baselines. A similar trend can also be observed in the YouTube dataset. Similarly, the median value of CMI on texts generated by {\modelname} is $0.27$ and $0.16$, respectively, for the Twitter and YouTube datasets, which are significantly lower than the median CMI achieved by other baselines ($0.29$ and $0.28$, respectively). A lower median CMI indicates that the texts generated by \modelname\ are more monolingual at a corpus level, acknowledging the population level trend shown in Figure~\ref{fig:user_data}. 

Figure~\ref{fig:word_analysis}(a) shows the distribution of top Hindi verbs and nouns from the Twitter dataset. Being more inclined towards monolingual, {\modelname} assigns more probability to these Hindi words, irrespective of the parts of speech. Figure~\ref{fig:word_analysis}(b) shows the distribution of top Hindi verbs and nouns from the Twitter dataset for different ablations of {\modelname}. Interestingly, {\modelname} with the alignment module can replicate the word distribution observed in the real dataset. On the other hand, without the alignment module, the generative model could hallucinate and unrealistically use common phrases in incorrect contexts. This highlights the effectiveness of the alignment module in recalibrating output tokens and generating semantically meaningful texts. Although the dimension of the alignment matrix is $|V| \times |V|$, with $|V|$ being the decoder vocabulary size, the learnable parameters are of order $O(d^2)$, where $d$ is the hidden dimension. The total number of additional parameters introduced by the alignment module is $0.025\%$ of the entire network, which is insignificant compared to other modules. 

\begin{figure*}[!t]
\centering
\subfloat[][]{\includegraphics[scale=0.2]{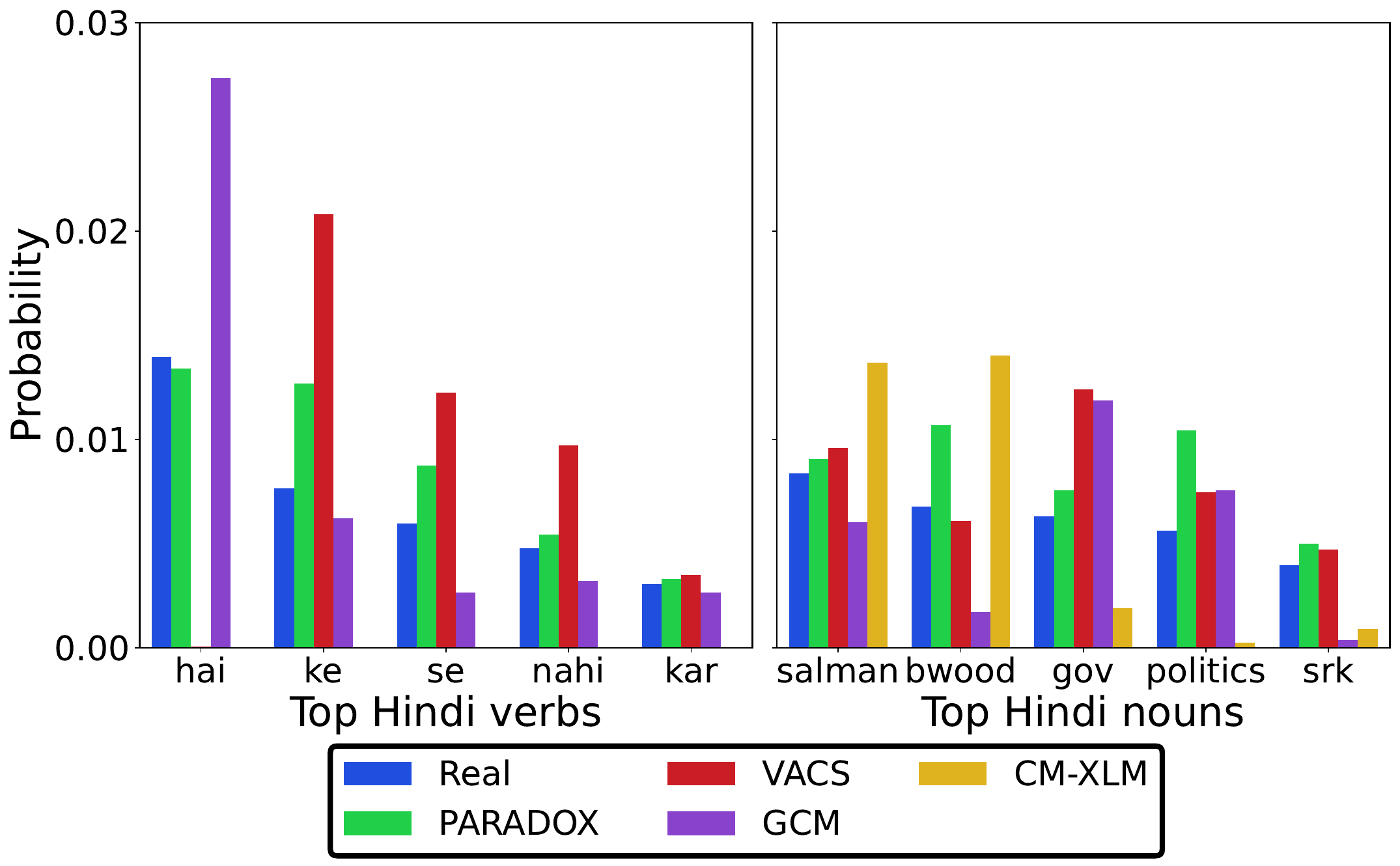}}
\quad
\subfloat[][]{\includegraphics[scale=0.2]{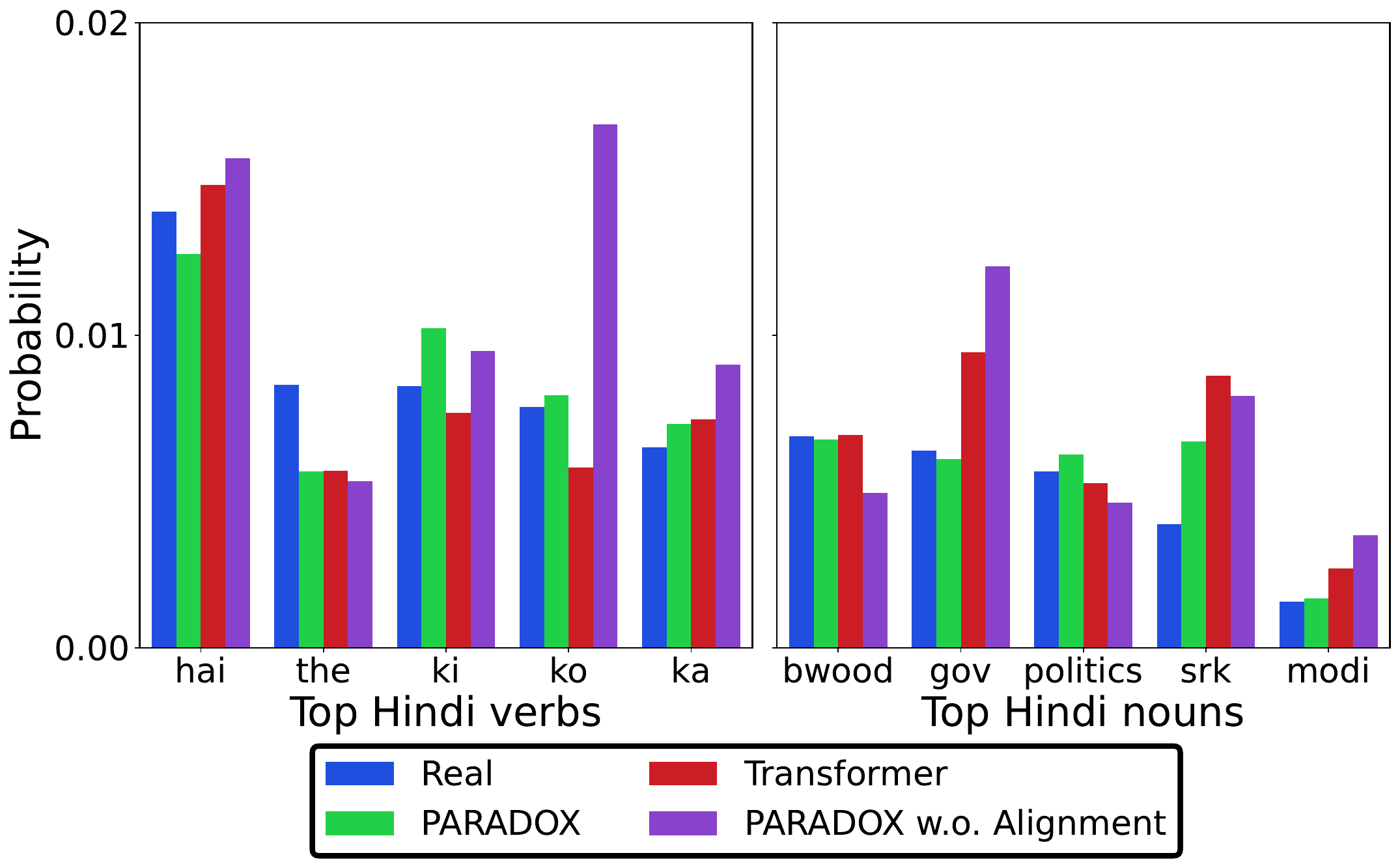}}
\caption{Distribution of top Hindi verbs and nouns for Tweets. w.o. alignment is the ablation of {\modelname} without the alignment module.}
\label{fig:word_analysis}
\end{figure*}

We further analyze a few sample texts generated by the generative models. We report examples of texts generated by {\modelname} and the Transformer, along with the average semantic coherence and linguistic quality scores annotated by the annotators in Table~\ref{tab:few_more_examples} of Appendix~\ref{sec:appx_analysis}. 
Transformer usually picks top nouns in the corpus and generates texts around them without considering the syntax of the code-mixed language, resulting in incoherent texts in many cases. On the other hand, {\modelname} preserves the grammar of code-mixed texts with a more human-like switching pattern. It shows that {\modelname} maintains the grammar of the base language (Hindi in this case), attributing to a more coherent and reliable generation. The examples highlight the key differences between the texts generated by persona-based {\modelname} and non-persona-based Transformer models regarding text quality. Table~\ref{tab:examples} further shows the personalized generation by our model. With the same prompt (\textit{e.g.,} `\textit{salman}' in the first example), {\modelname} can understand different personas of users and can generate texts suited for different users. The high similarity between the historical average of the user CMIs and the generated CMIs indicates the model's ability to understand the linguistic preferences of users. \\

Table~\ref{tab:llama_examples}  of Appendix~\ref{sec:appx_analysis} highlights code-mixed texts generated by the Llama 2 model with zero-shot and 1-shot. {\modelname} exhibits better capability in mimicking the code-mixing linguistic traits than Llama. For user ID $2226$, who has had more monolingual usage in the past (average CMI $0.04$), the text generated by Llama is more code-mixed than monolingual. Similarly, for user ID $3$, the Llama model reverses the linguistic preference of the user while generating the code-mixed texts. Albeit demonstrating superior performance across various natural language understanding and reasoning tasks with zero and few-shot in-context learning, pre-trained large language models such as Llama fail to understand the linguistic complexities of informal languages such as code-mixed Hinglish. It is imperative to notice that the Llama model not only fails to capture the historical linguistic preferences of users but also fails to impersonate the semantic structure of code-mixed texts. On the other hand, {\modelname} demonstrates better code-mixed language understanding capabilities, captures the linguistic preferences of the user from their historical utterances and preserves the information for future generations. This highlights the effectiveness of personification in code-mixed text generation and the importance of building more robust language understanding models for understanding the linguistic nitty-gritty of low-resource languages.

\newcommand{\examples}{
\rotatebox{0}{\scalebox{0.9}{
\begin{tabular}{c| p{20em}| c | c}
\hline
User ID & Generated Text & Historical Avg. CMI & Generated CMI\\
\hline
\multirow{2}{*}{3} & \textbf{CM:} \textcolor{blue}{salman} ji please aapse milna hai & \multirow{2}{*}{0.28} & \multirow{2}{*}{0.33} \\
& \textbf{Eng:} Salman, I want to meet you, please & \\
\hdashline
\multirow{2}{*}{264} & \textbf{CM:} \textcolor{blue}{salman} ka fan ho na. & \multirow{2}{*}{0.50} & \multirow{2}{*}{0.40}\\
& \textbf{Eng:} You are Salman's fan, right? & \\
\hline
\multirow{2}{*}{762} & \textbf{CM:} \textcolor{blue}{ye} politics kr raha kya kr rahi h or truth show & \multirow{2}{*}{0.15} & \multirow{2}{*}{0.30}\\
& \textbf{Eng:} Is this politics, or truth show & \\
\hdashline
\multirow{2}{*}{2226} & \textbf{CM:} \textcolor{blue}{ye} fb ho jaye bhaijaan & \multirow{2}{*}{0.04} & \multirow{2}{*}{0.00}\\
& \textbf{Eng:} This has became fb, brother \\
\hline
\end{tabular}
\label{tab:examples}
}}}

\begin{table*}[!t]
 \centering
  \examples
  \caption{Example of different prompted (prompts highlighted with \textcolor{blue}{blue}) generation for different users. Different CMI indicates the difference in prompted generation based on the user persona.}%
  \label{tab:examples}%
\end{table*}

\section{Conclusion}
This paper described a personalized code-mixed generation model for generating human-alike code-mixed texts. We highlighted the need for a personalized generation under the pretext of code-mixing. Toward this, we devised a novel persona-aware encoder-decoder model coupled with a novel alignment module for generating more realistic and coherent Hindi-English code-mixed texts, the first attempt toward personalized code-mixed generation. Empirical analyses would benefit the research community in developing robust and reliable language models for low-resource languages. \color{black}Although our current work explored \modelname\ for Hindi-English generation, the proposed generative framework can be further extended to generate code-mixed texts in other personalized code-mixing and derived languages, such as Spanish-English and Chinese-English. \color{black} Although our empirical study has shown the effectiveness of persona-attributed text generation, currently {\modelname} captures only contextual persona, ignoring other explicit factors. Not only does this restrict our model in cold-start generation (text generation for new users without any history), but it also fails to consider the co-association among users. In conversational settings, particularly, this can be deemed essential. Another limitation of {\modelname} is the inability to determine the temporal evolution of a user's persona driven by external factors. {\modelname} captures the user persona and its evolution solely from contextual information. At the same time, a user's linguistic preferences can also be driven by other external socio-demographic and economic factors varying over time, which our model currently undermines.

\subsubsection*{Broader Impact Statement}

Our work highlights the need for personalized generation models for conversational languages like code-mixing. We release our curated datasets to encourage research on personalized code-mixed text generation. Persona-aware code-mixed generation models can aid in building data-driven solutions in low-resource languages and can be expanded to broader demographics. \color{black} We do not collect user-specific features or determinants that could reveal user-specific sensitive information. We remove all the user IDs, mentions and delimiters from the user texts to maintain user privacy. Moreover, no user-specific attributes are used to train the generative model except the texts written by the users. \color{black} Although we do not anticipate any immediate negative impact of our work, over-personalization can lead to targeted spamming and negative misuse of user persona. We ask the researchers to be aware of the potential misuse and use the shared artefacts judiciously to prevent unwarranted events.


\bibliography{paper}
\bibliographystyle{tmlr}

\appendix
\section{Experimental Setup}
\subsection{Evaluation Metrics}
We compute Pearson correlation (c.f. Figure~\ref{fig:heatmap}) between the proposed metrics to understand their relationships. We observe strong positive correlations among the metrics -- CM BLEU, CM Rouge-1 and CM Rouge-L. Similarly, all of these metrics have a strong negative correlation with perplexity. This highlights the strong linear associations between the evaluation metrics that measure the semantical behavior of the code-switching patterns. On the other hand, the correlations between the CM KS measure and other metrics are meagre, indicating that the linguistic preferences of users have very minimal impact on the switching patterns. In other words, the switching patterns of a user who prefers multilingual could be more erratic than those of a user who prefers monolinguals. 
\label{sec:appx_eval_metric}

\begin{figure}[!ht]
\centering
{\includegraphics[scale=0.3]{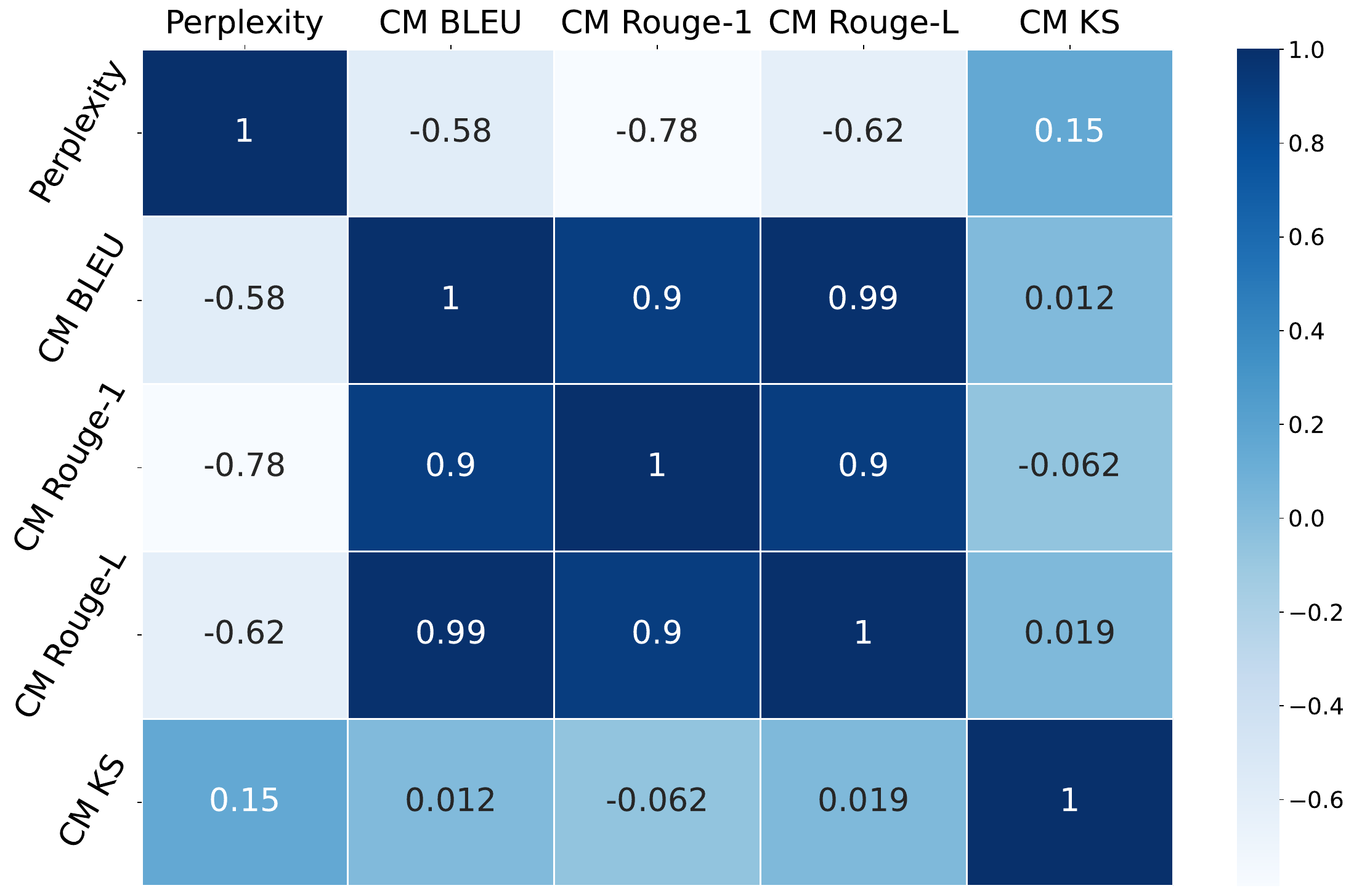}}
\caption{Pearson correlation between different evaluation measures on the validation dataset.}
\label{fig:heatmap}
\end{figure}

\subsection{Training Details}
\label{sec:appx_training_details}
For all the models across all the experiments, we use a maximum text length of $40$. {\modelname} consists of six encoder and decoder layers, with hidden sizes of $768$ in all the layers. For multi-headed FAME and masked multi-headed FAME blocks, we use a total of eight heads with $Dropout$ probabilities set as $0.1$. The total number of parameters is $296$M. We use six encoder and six decoder layers in the Transformer model, with eight heads in each multi-headed attention block. For training {\modelname}, We set the persona encoding variational weight $\lambda = 0.5$. All the models are trained for $50$ epochs with an early stopping condition on validation loss with the patience of $10$. We set $batch\_size = 4$ in all experiments during training and validation. We use Adam optimizer with a learning rate of $4e-4$ and $\beta_1 = 0.9, \beta_2 = 0.98$ for both {\modelname} and Transformer. We fine-tune the MuRIL and BLOOMZ models on autoregressive language modeling tasks for $10$ epochs with learning rates $3e-5$ and $3e-6$, respectively. The Llama-2 baseline is used in zero-shot and 1-shot settings with the prompt shown in Figure~\ref{fig:llama_instruct}. We use one Tesla P100 and one Tesla V100 GPU to run all our experiments. For {\modelname}, each training and validation iteration takes $\sim 0.18$ and $\sim 0.12$ seconds, respectively. ~\cite{strubell-etal-2019-energy} proposed estimation of power usage and carbon emission behind running deep learning experiments. Following those guidelines, we estimate a total power usage of $23.56$ kWh and an equivalent CO2 emission of $22.46$ pounds.

\begin{figure}
\centering
\includegraphics[scale=0.4]{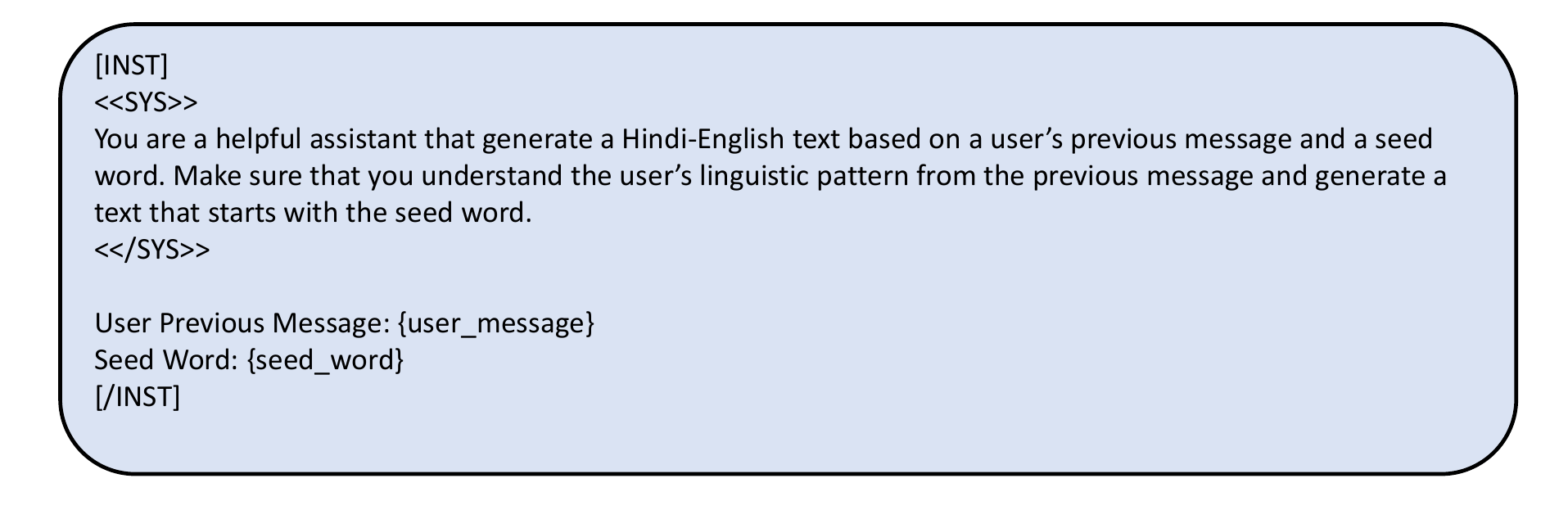}
\caption{Prompt used with the Llama 2 model for Hindi-English code-mixed text generation.}
\label{fig:llama_instruct}
\end{figure}

\section{Analysis of Code-Mixed Generation}
\label{sec:appx_analysis}
We highlight few examples of texts generated by \modelname\ and the vanilla Transformer model in Table~\ref{tab:few_more_examples}, along with their semantic coherence and linguistic quality scores annotated by the annotators. As highlighted in Table~\ref{tab:human_evaluation}, the semantic coherence of the texts generated by \modelname\ is much higher than the non-persona-based counterpart. Moreover, texts generated by \modelname\ are grammatically more valid than those generated by the Transformer model. This analysis highlights the importance of incorporating persona for an enhanced understanding of conversational code-mixed languages.

Table~\ref{tab:llama_examples} highlights the texts generated by the persona-based \modelname\ and the Llama model for two different users. The margin between the historical average CMI and the generated CMI highlights our method's superiority in capturing users' linguistic patterns. A large pre-trained language model such as Llama, even with few-shot in-context learning, may not be superior to a persona-based small language model fine-tuned on low-resource code-mixed texts. 

\begin{table*}[t!]
\centering
\rotatebox{0}{\scalebox{0.9}{
\begin{tabular}{c| p{16em}| c | c}
\hline
Model & Generated Text & Semantic Coherence & Linguistic Quality\\
\hline
\multirow{4}{*}{\modelname} & \textbf{CM:} are bhai apne! & 3.70 & 3.60\\
& \textbf{Eng:} Hey my brother! & & \\
\cdashline{2-4}
& \textbf{CM:} bhai bahut thik ho gaya & 4.27 & 4.27 \\
& \textbf{Eng:} brother is very well & & \\
\hline
\multirow{4}{*}{Transformer} & \textbf{CM:} rockstar walo se samjha media........ & 3.07 & 3.57\\
& \textbf{Eng:} Media understood by rockstars & & \\
\cdashline{2-4}
& \textbf{CM:} boy likha hai thanks bhai & 3.57 & 3.53 \\
& \textbf{Eng:} Boy has written thanks brother & & \\
\hline
\end{tabular}}}
\caption{Examples of code-mixed texts generated by {\modelname} and Transformer with human annotated average Semantic coherence and linguistic quality scores.} \label{tab:few_more_examples}
\end{table*}
\newcommand{\llamaexamples}{
\rotatebox{0}{\scalebox{0.8}{
\begin{tabular}{c| c | p{22em}| c | c}
\hline
User ID & Model & Generated Text & Historical Avg. CMI & Generated CMI\\
\hline
\multirow{6}{*}{3} & \multirow{2}{*}{\modelname} & \textbf{CM:} salman ji please aapse milna hai & \multirow{2}{*}{0.28} & \multirow{2}{*}{0.33} \\
& & \textbf{Eng:} Salman, I want to meet you, please & & \\
\hdashline
& \multirow{2}{*}{Llama zero-shot} & \textbf{CM:} Salman ne kaha tha flash me jeetega apka siddhant & \multirow{2}{*}{0.28} & \multirow{2}{*}{0.22} \\
& & \textbf{Eng:} Salman had said flushme jeetega apna principle & & \\
\hdashline
& \multirow{2}{*}{Llama 1-shot} & \textbf{CM:} Salman bhai ne kaha, Flush mein jitega apna side, main toh bus se nikal raha hoon! & \multirow{2}{*}{0.28} & \multirow{2}{*}{0.11} \\
& & \textbf{Eng:} Salman had said flushme jeetega apna side, I am just leaving in a bus & & \\
\hline

\multirow{6}{*}{2226} & \multirow{2}{*}{\modelname} & \textbf{CM:} ye fb ho jaye bhaijaan & \multirow{2}{*}{0.04} & \multirow{2}{*}{0.00} \\
& & \textbf{Eng:} This has became fb, brother & & \\
\hdashline
& \multirow{2}{*}{Llama zero-shot} & \textbf{CM:} ye superb jabardast bahut khub, bhadaai ho & \multirow{2}{*}{0.04} & \multirow{2}{*}{0.37} \\
& & \textbf{Eng:} This is superb, amazing, and congratulations & & \\
\hdashline
& \multirow{2}{*}{Llama 1-shot} & \textbf{CM:} ye superb jabardast bahut khub, bhadaai ho & \multirow{2}{*}{0.04} & \multirow{2}{*}{0.37} \\
& & \textbf{Eng:} This is superb, amazing, and congratulations & & \\
\hline

\end{tabular}
\label{tab:llama_examples}
}}}

\begin{table*}[t]
 \centering
  \llamaexamples
  \caption{Example of texts generated by {\modelname} and Llama 2 models.}%
  \label{tab:llama_examples}%
\end{table*}

\end{document}